\documentclass[cameraready]{Interspeech}

\newcommand{\bn}[2]{\textit{#2}}

\usepackage{multirow}

\usetikzlibrary{arrows.meta,positioning}

\ifcameraready\else
  \renewcommand\anonname{Anonymous submission}
\fi

\title{BanglaTurn: A Benchmark and Whisper-Based Model for End-of-Turn Detection in Bangla Speech}

\author[affiliation={1}, orcid=0009-0000-2297-6037]{Mizbaul Haque}{Maruf}

\address{
    $^1$ Vivasoft Limited, Dhaka, Bangladesh
}

\email{mizbaul.haque@vivasoftltd.com}


\begin{document}

\maketitle

\begin{abstract}
    This paper presents BanglaTurn, a corpus for end-of-turn detection in Bangla conversational speech, and a model trained on it. The corpus holds 35,374 samples of 3 to 15 s of podcast speech, labelled for turn state by combining speaker diarization with an LLM pass, with every label then checked by a human annotator. The model pairs a Whisper encoder with task-specific classification heads. On a class-balanced test set drawn from a held-out podcast, it reaches 84.33\% accuracy (95\% CI 80.3 to 88.1) against 69.28\% for the Smart-Turn v3 baseline, and lowers the false negative rate from 51.57\% to 7.55\% at the cost of a higher false positive rate. We report what encoder layer fine-tuning, multi-scale pooling and INT8 quantization each contribute, and latency stays within 165 to 191 ms end to end on CPU.
\end{abstract}

\section{Introduction}

Turn-taking governs how speakers coordinate the exchange of speaking turns \cite{sacks1974simplest}. Detecting when a speaker has finished, the task of end-of-turn detection, matters for conversational AI systems \cite{raux2009optimizing}: get it wrong and the system either cuts the user off or leaves an awkward gap, and users notice both in voice assistants and spoken dialogue applications. We use \emph{turn detection} as shorthand for this task throughout.

Early systems relied on silence thresholds \cite{ferrer2002speaker}, which cannot separate a genuine turn ending from a pause in the middle of an utterance. Later work modelled acoustic and prosodic cues with neural networks \cite{skantze2017towards,roddy2018investigating}, and self-supervised representations from Wav2Vec 2.0 \cite{baevski2020wav2vec}, HuBERT \cite{hsu2021hubert} and Whisper \cite{radford2022robust} made transfer learning practical for the task. Low-resource languages have seen little of this progress. Bangla has more than 230 million speakers but no dedicated turn detection dataset, and multilingual models such as Smart-Turn v3 do poorly on it because Bangla conversational speech is thinly represented in their training data.

This paper makes two contributions. The first is BanglaTurn, a Bangla turn detection corpus of 35,374 annotated samples. The second is a Whisper encoder-based model that reaches 84.33\% accuracy, 15 points above the baseline, at speeds usable in a live system.

\begin{figure}[t]
    \centering
    \includegraphics[width=\linewidth]{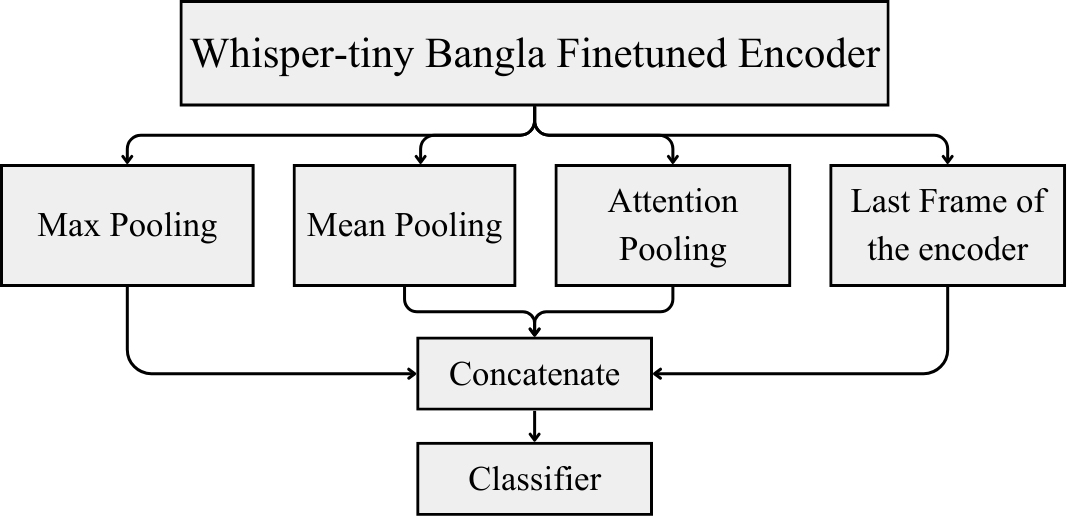}
    \caption{Our proposed model architecture with Whisper-tiny encoder finetuned in Bangla, multi-scale pooling and a classifier.}
    \label{fig:architecture}
\end{figure}

\section{Related Work}

\subsection{Turn detection systems}

Early turn detection systems combined silence thresholds with prosodic features such as pitch contours and energy patterns \cite{ferrer2002speaker,gravano2011turncues}. Statistical models (HMMs, CRFs) then captured temporal dependencies across acoustic and linguistic features. Neural approaches followed, first with continuous LSTM-based prediction \cite{skantze2017towards}, then with models that fused acoustic, prosodic and lexical cues \cite{roddy2018investigating}. More recently, TurnGPT \cite{ekstedt2020turngpt} predicts turn-taking events from text using language-model pretraining, while Voice Activity Projection \cite{ekstedt2022voice} learns them from self-supervised speech representations. Other work targets latency-aware prediction for real-time systems \cite{inoue2022latency} and multi-party conversation \cite{masumura2018neural}. Prosody remains central throughout: Ward and Vega \cite{ward2019prosodic} show that phrase-final words signal completion through falling pitch, lengthening and reduced intensity. Work presented at O-COCOSDA has studied the same cues from the corpus side. Furukawa et al.\ \cite{furukawa2011multimodal} built a multimodal corpus for modelling turn management in multi-party conversation, and Ishimoto and Enomoto \cite{ishimoto2016endofutterance} showed that final pitch lowering shapes how listeners perceive the end of an utterance in spontaneous Japanese.

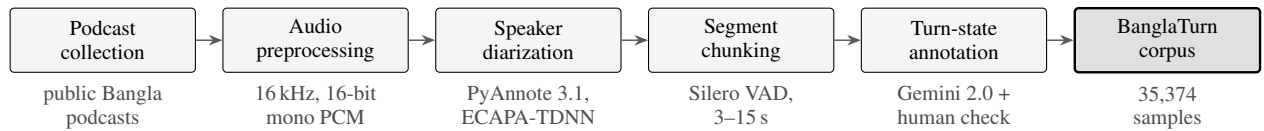
\begin{figure*}[t]
    \centering
    \begin{tikzpicture}[
        stage/.style={draw=black, line width=0.4pt, rounded corners=1.5pt,
                      fill=black!4, minimum width=24.5mm, minimum height=8.5mm,
                      align=center, inner sep=1.5pt, font=\footnotesize},
        output/.style={stage, fill=black!12, line width=0.9pt},
        note/.style={align=center, font=\footnotesize, text=black!70, inner sep=0pt},
        flow/.style={-{Stealth[length=1.8mm,width=1.4mm]}, line width=0.55pt,
                     draw=black!65},
      ]
        \node[stage]                     (s1) {Podcast\\collection};
        \node[stage,  right=3.5mm of s1] (s2) {Audio\\preprocessing};
        \node[stage,  right=3.5mm of s2] (s3) {Speaker\\diarization};
        \node[stage,  right=3.5mm of s3] (s4) {Segment\\chunking};
        \node[stage,  right=3.5mm of s4] (s5) {Turn-state\\annotation};
        \node[output, right=3.5mm of s5] (s6) {BanglaTurn\\corpus};

        \foreach \a/\b in {s1/s2, s2/s3, s3/s4, s4/s5, s5/s6}{
            \draw[flow] (\a) -- (\b);
        }

        \node[note, below=1.6mm of s1] {public Bangla\\podcasts};
        \node[note, below=1.6mm of s2] {16\,kHz, 16-bit\\mono PCM};
        \node[note, below=1.6mm of s3] {PyAnnote 3.1,\\ECAPA-TDNN};
        \node[note, below=1.6mm of s4] {Silero VAD,\\3--15\,s};
        \node[note, below=1.6mm of s5] {Gemini 2.0 +\\human check};
        \node[note, below=1.6mm of s6] {35,374\\samples};
    \end{tikzpicture}
    \caption{Dataset creation workflow. Each box names a stage, with the tool or setting it uses beneath it: public podcasts are normalised to \SI{16}{\kilo\hertz} mono, split into speaker turns by PyAnnote, cut into \SIrange[range-phrase={--},range-units=single]{3}{15}{\second} chunks by Silero VAD, labelled for turn state by combining the diarization with Gemini 2.0 Flash Lite, with every label checked by a human annotator, and collected into the 35,374-sample BanglaTurn corpus.}
    \label{fig:workflow}
\end{figure*}

\subsection{Speech representations}

Self-supervised models learn transferable representations from unlabelled audio. Wav2Vec 2.0 \cite{baevski2020wav2vec} uses contrastive learning over quantized representations, HuBERT \cite{hsu2021hubert} predicts masked discrete hidden units, and XLSR \cite{conneau2020unsupervised} extends the approach to low-resource languages through cross-lingual pretraining.

Whisper \cite{radford2022robust} is trained on 680,000 hours of weakly supervised data covering 96 languages, Bangla among them, which makes its encoder a reasonable starting point for transfer \cite{howard2018universal}. Fine-tuning the top encoder layers while freezing the lower ones trades adaptation against generalization \cite{yosinski2014transferable}. We are not aware of work that adapts these models to turn detection in a low-resource language.

\subsection{Bangla speech processing}

Bangla is underrepresented relative to high-resource languages. Mozilla Common Voice \cite{ardila-etal-2020-common} and Kjartansson et al.\ \cite{kjartansson2018crowdsourced} provide crowdsourced Bangladeshi Bengali speech, and BanglaBERT \cite{islam2022banglabert} set benchmarks for Bangla language understanding. Bengali speech resources also have a history at O-COCOSDA, from a read-speech corpus for continuous speech recognition \cite{das2011bengali} to a prosodically annotated Bengali and Assamese audiobook corpus for sentence boundary detection \cite{chowdhury2025prosodically}. Sentence boundaries in read speech are related to turn ends in conversation but are not the same thing, since a conversational speaker can pause at a sentence boundary and still hold the floor. Turn detection has nothing comparable. There is no public Bangla dataset or benchmark for it, and multilingual models degrade on Bangla because of its prosody and its small share of training data. We built BanglaTurn to fill that gap.

\section{BanglaTurn Detection Dataset}

Conversational Bangla audio is scarce, so we built a pipeline that collects it, separates the speakers, cuts the result into usable segments, and annotates them with model assistance.

\subsection{Dataset creation}

Figure~\ref{fig:workflow} shows the pipeline. The audio comes from Bangla-language conversational podcasts that are publicly viewable on YouTube. We chose podcasts because they are conversational, reasonably clean, and cover a range of topics. Section~\ref{sec:ethics} describes the gated terms under which we release this audio. Audio was normalized to \SI{16}{\kilo\hertz}, 16-bit mono PCM. Speaker boundaries come from PyAnnote Speaker Diarization 3.1 \cite{bredin2020pyannote}, which combines voice activity detection, ECAPA-TDNN embeddings \cite{snyder2018x} and agglomerative clustering. We merge adjacent segments from the same speaker and discard anything under \SI{2.0}{\second}. Where the speakers are known in advance, we identify them automatically by verification against reference samples.

Diarization returns segments of arbitrary length, but the model takes a fixed-duration input, so we chunk with Silero VAD under two constraints: each chunk runs \SIrange{3.0}{15.0}{\second}, with at most \SI{2.0}{\second} of trailing silence.

\subsection{Label scheme}
\label{sec:labels}

Each sample carries one target label and two auxiliary flags. The target is binary. An \textit{endpoint} is a clip that ends with a genuine turn completion, where the speaker yields the floor (e.g.\ ``\bn{আমি কাল বাজারে গিয়েছিলাম।}{ami kal bajare giyechilam.}'', ``I went to the market yesterday.''). A \textit{non-endpoint} is a clip after which the same speaker carries on. The model is trained and evaluated on this label alone.

The two flags mark hesitation sounds and discourse fillers, such as ``\bn{উম}{um}'' or ``\bn{মানে}{mane}'' (``I mean''). They are not classes. \textit{Mid-filler} marks a filler inside the clip (e.g.\ ``\bn{আমি... আচ্ছা... ভাবছিলাম যে আমরা এটা করতে পারি।}{ami... accha... bhabchilam je amra eta korte pari.}'', ``I... well... was thinking that we can do this.''). \textit{End-filler} marks a filler in the last one to three words (e.g.\ ``\bn{আমি কাল বাজারে গিয়েছিলাম... উহ...}{ami kal bajare giyechilam... uh...}'', ``I went to the market yesterday... uh...''). The flags are independent of each other and of the target. A speaker can trail off with a filler and then yield the floor, or finish a filler-free sentence and keep going. In the test set, 248 of the 319 clips carry a mid-filler and 104 an end-filler, and both occur with either label. We keep the flags because fillers are where a pause is most easily mistaken for a turn end, and a system that makes that mistake either interrupts the speaker or stalls.

\subsection{Annotation protocol}
\label{sec:annotation}

Labels come from two automatic sources, which a human annotator then reconciled. The first is diarization. The last chunk of a speaker's turn before a different speaker begins is proposed as an endpoint, and every other chunk as a non-endpoint. This rule is noisy, because diarization errors, overlapping speech and backchannels all create spurious speaker changes. The second is an LLM pass with Google's Gemini 2.0 Flash Lite, following the LLM-assisted annotation approach of Gilardi et al.\ \cite{gilardi2023chatgpt}. For each clip, a fixed prompt asks for a Bangla transcription, the two filler flags, and one of four turn states, each defined with examples: \textit{Complete} (intent fully expressed), \textit{Incomplete} (a pause inside an unfinished thought), \textit{Backchannel} (a brief listener response) and \textit{Wait} (a request to pause or stop). Complete and Wait map to endpoint, Incomplete and Backchannel to non-endpoint, and only the binary label is kept.

A single annotator, the author, a native Bangla speaker, then listened to every clip alongside both proposals and the transcription. Where the proposals agreed, the annotator confirmed or corrected the label. Where they disagreed, the annotator decided by listening to the clip alone, without the surrounding audio. Filler flags and transcriptions were checked in the same pass.

This protocol has two gaps. Because one person checked every label, we cannot report inter-annotator agreement. Because corrections were made in place without a log, we cannot report how often the human check changed the automatic labels. Section~\ref{sec:discussion} considers what this means for the results.

\subsection{Dataset details}

BanglaTurn holds 35,374 samples split into training (31,549), validation (3,506) and test (319) sets, as detailed in Table~\ref{tab:dataset_stats}. Training and validation keep the natural class distribution of roughly 80\% endpoints, while the test set is balanced at 49.84\% endpoints. The test set is small, but it is not a random in-domain split. It consists of handpicked, class-balanced and deliberately difficult examples taken from a separate podcast that was held out entirely from training and validation. The same annotator who checked the labels selected these examples. The 319 clips run \SIrange[range-phrase={ to },range-units=single]{3.0}{15.0}{\second} and total 51.6 minutes. Because the test set comes from a different speaker and a different distribution, it tests generalization rather than in-domain memorization, which we consider a harder and more informative check than a larger split drawn from the same recordings. Its small size does make every figure uncertain, so we report confidence intervals (Section~\ref{sec:setup}).

\begin{table}[t]
  \caption{BanglaTurn dataset statistics. Endpoint and non-endpoint partition each split.}
  \label{tab:dataset_stats}
  \centering
  \begin{tabular}{lccc}
    \toprule
    \textbf{Metric} & \textbf{Training} & \textbf{Validation} & \textbf{Test} \\
    \midrule
    Total samples & 31,549 & 3,506 & 319 \\
    Endpoint      & 25,276 & 2,829 & 159 \\
    Non-endpoint  & 6,273  & 677   & 160 \\
    \bottomrule
  \end{tabular}
\end{table}

\begin{table*}[t]
  \caption{Performance comparison on the BanglaTurn test set (all metrics in \%, endpoint as the positive class). Brackets give 95\% bootstrap confidence intervals. FPR is computed over the 160 non-endpoint clips and FNR over the 159 endpoint clips. $^\dagger$Fitted on the test podcast itself by 10-fold cross-validation, and therefore optimistic.}
  \label{tab:main_results}
  \centering
  \setlength{\tabcolsep}{4.0pt}
  \begin{tabular}{lcccccc}
    \toprule
    \textbf{Model} & \textbf{Acc} & \textbf{FPR} & \textbf{FNR} & \textbf{Prec} & \textbf{Rec} & \textbf{F1} \\
    \midrule
    Silence threshold$^\dagger$ & 49.22 [43.9, 54.9] & 11.88 & 89.94 & 45.71 & 10.06 & 16.49 [9.6, 23.5] \\
    Prosody + silence LR$^\dagger$ & 62.07 [56.7, 67.4] & 50.00 & 25.79 & 59.60 & 74.21 & 66.11 [60.2, 71.4] \\
    Smart-Turn v3 & 69.28 [63.9, 74.3] & \textbf{10.00} & 51.57 & \textbf{82.80} & 48.43 & 61.11 [53.7, 67.7] \\
    \textbf{Ours} & \textbf{84.33} [80.3, 88.1] & 23.75 & \textbf{7.55} & 79.46 & \textbf{92.45} & \textbf{85.47} [81.2, 89.3] \\
    \bottomrule
  \end{tabular}
\end{table*}

\begin{table*}[t]
  \caption{Ablation study comparing different model variants on the BanglaTurn test set. MSP denotes Multi-Scale Pooling. FPR and FNR are computed over the non-endpoint and endpoint clips respectively. With 319 test clips, the 95\% confidence interval on each accuracy spans about $\pm$4 points, wider than most gaps between rows.}
  \label{tab:ablation}
  \centering
  \begin{tabular}{lccccccc}
    \toprule
    \textbf{Encoder + Components} & \textbf{Unfrozen} & \textbf{Acc} & \textbf{FPR} & \textbf{FNR} & \textbf{Prec} & \textbf{Rec} & \textbf{F1} \\
    & \textbf{layers} & \textbf{(\%)} & \textbf{(\%)} & \textbf{(\%)} & \textbf{(\%)} & \textbf{(\%)} & \textbf{(\%)} \\
    \midrule
    Whisper-tiny & 0 & 83.07 & 21.88 & 11.95 & 80.00 & 88.05 & 83.83 \\
    Whisper-tiny & 2 & 81.19 & 29.38 & 8.18 & 75.65 & 91.82 & 82.95 \\
    Whisper-tiny & 4 & 83.07 & 27.50 & 6.29 & 77.20 & 93.71 & 84.66 \\
    Whisper-tiny (Bangla) & 0 & 84.01 & \textbf{20.62} & 11.32 & \textbf{81.03} & 88.68 & 84.68 \\
    Whisper-tiny (Bangla) & 2 & 81.19 & 27.50 & 10.06 & 76.47 & 89.94 & 82.66 \\
    Whisper-tiny (Bangla) & 4 & 81.50 & 29.38 & 7.55 & 75.77 & 92.45 & 83.29 \\
    Whisper-tiny (Bangla) + MSP & 0 & 83.07 & 24.38 & 9.43 & 78.69 & 90.57 & 84.21 \\
    \textbf{Whisper-tiny (Bangla) + MSP} & \textbf{2} & \textbf{84.33} & 23.75 & 7.55 & 79.46 & 92.45 & \textbf{85.47} \\
    Whisper-tiny (Bangla) + MSP & 4 & 83.07 & 28.12 & \textbf{5.66} & 76.92 & \textbf{94.34} & 84.75 \\
    \bottomrule
  \end{tabular}
\end{table*}

\begin{table}[t]
  \caption{Confusion matrices on the BanglaTurn test set (159 endpoint and 160 non-endpoint clips). Rows give the reference label, columns the prediction. $^\dagger$Fitted on the test podcast by 10-fold cross-validation.}
  \label{tab:confusion}
  \centering
  \setlength{\tabcolsep}{3.5pt}
  \renewcommand{\arraystretch}{1.1}  
  \begin{tabular}{p{2.1cm}lcc}
    \toprule
    & & \multicolumn{2}{c}{\textbf{Predicted}} \\
    \cmidrule(lr){3-4}
    \textbf{Model} & \textbf{Reference} & \textbf{Endpoint} & \textbf{Non-endpoint} \\
    \midrule
    \multirow{2}{2.1cm}{\raggedright Silence threshold$^\dagger$} & Endpoint & 16 & 143 \\
     & Non-endpoint & 19 & 141 \\
    \midrule
    \multirow{2}{2.1cm}{\raggedright Prosody + silence LR$^\dagger$} & Endpoint & 118 & 41 \\
     & Non-endpoint & 80 & 80 \\
    \midrule
    \multirow{2}{2.1cm}{\raggedright Smart-Turn v3} & Endpoint & 77 & 82 \\
     & Non-endpoint & 16 & 144 \\
    \midrule
    \multirow{2}{2.1cm}{\raggedright Ours} & Endpoint & 147 & 12 \\
     & Non-endpoint & 38 & 122 \\
    \bottomrule
  \end{tabular}
\end{table}

\begin{table}[t]
  \caption{Impact of INT8 quantization on model size, inference latency, and accuracy.}
  \label{tab:quantization}
  \centering
  \footnotesize  
  \setlength{\tabcolsep}{2.6pt}
  \begin{tabular}{lcccc}
    \toprule
    \textbf{Encoder + Components} & \textbf{Quant.} & \textbf{Acc} & \textbf{Size} & \textbf{Lat.} \\
    & & \textbf{(\%)} & \textbf{(MB)} & \textbf{(ms)} \\
    \midrule
    Whisper-tiny & FP32 & 83.07 & 148.2 & 182.92 \\
    Whisper-tiny & INT8 & 82.76 & 39.4 & 164.98 \\
    \midrule
    Whisper-tiny (Bangla) & FP32 & 84.01 & 148.2 & 180.04 \\
    Whisper-tiny (Bangla) & INT8 & 79.31 & 39.4 & 169.96 \\
    \midrule
    Whisper-tiny (BN) + MSP & FP32 & 84.33 & 149.1 & 191.00 \\
    Whisper-tiny (BN) + MSP & INT8 & 83.07 & 39.8 & 165.95 \\
    \bottomrule
  \end{tabular}
\end{table}

\section{Model Architecture}

Our model reuses Whisper's encoder for binary turn detection, as Figure~\ref{fig:architecture} shows. The task does not call for transcription, only for the acoustic, prosodic and temporal structure of the audio, and the encoder already represents that, so we drop the decoder.

\subsection{Encoder}

We use a Whisper Tiny encoder fine-tuned on the Mozilla Common Voice 11 Bangla dataset \cite{ardila-etal-2020-common}. Its input is an 80-channel log-Mel spectrogram covering 8 seconds of audio.

\subsection{Multi-scale pooling}

Turn detection depends on local acoustic events, such as the final phoneme and the intonation contour, and on the shape of the utterance as a whole. To capture both, we concatenate four pooled views of the encoder output: max pooling for prominent features, mean pooling for the overall representation, attention pooling with a learnable query that can settle on turn-relevant regions, and last-frame pooling for the final temporal state.

\subsection{Classification head}

The head is a stack of linear layers with layer normalization, GELU activations and dropout, ending in a single logit.

\section{Experimental Setup}
\label{sec:setup}

We compare against Smart-Turn v3 on the same BanglaTurn test split and ablate three variants: the stock Whisper-tiny encoder, trained on 99 languages including Bangla; the same encoder fine-tuned on Bangla; and that Bangla fine-tuned encoder with multi-scale pooling.

We also compare against two classical baselines that use no learned speech representation. The first is a silence threshold, which predicts an endpoint when the clip's trailing silence is at least $\tau$ seconds. Trailing silence is the time after the last frame within 35\,dB of the clip's peak energy. The second is a logistic regression over trailing silence and four prosodic features of the final 500\,ms of speech: F0 slope and F0 level relative to the clip median, both from pYIN \cite{mauch2014pyin}, energy slope, and the duration of the final speech segment, plus an indicator for clips with no final pitch estimate. We fit both on the test podcast itself by stratified 10-fold cross-validation, choosing $\tau$ or the regression weights on nine folds and predicting the tenth. The baselines thereby see the test speaker and recording conditions, which the neural models never do, so their scores are optimistic.

Training uses Binary Cross-Entropy (logits) loss with dynamic class weighting, which keeps the model from collapsing onto the majority class. We use AdamW \cite{loshchilov2019adamw} at learning rate 5e-5, weight decay 0.01 and gradient clipping at 1.0, for 4 epochs at batch size 16, with a 0.2 linear warmup ratio and cosine decay, on a single NVIDIA A40 (\SI{48}{\giga\byte}).

Two error rates matter most here. A false positive, predicting an endpoint while the speaker is still mid-turn, makes the system interrupt. A false negative makes it wait when it should respond. With endpoint as the positive class and a decision threshold of 0.5, we report the false positive rate FPR $= \mathrm{FP}/(\mathrm{FP}+\mathrm{TN})$ and the false negative rate FNR $= \mathrm{FN}/(\mathrm{FN}+\mathrm{TP})$ alongside accuracy, precision, recall and F1, together with inference latency and model size. Because the test set is small, we attach 95\% percentile bootstrap confidence intervals to accuracy and F1, computed from 10,000 resamples of the 319 test clips.

\section{Results and Analysis}

\subsection{Overall performance comparison}

Our model reaches 84.33\% accuracy where Smart-Turn v3 reaches 69.28\% (Table~\ref{tab:main_results}). The confidence intervals do not overlap. An unpaired two-proportion test on accuracy, which is conservative for two models scored on the same clips, gives $z = 4.50$, $p < 10^{-5}$. The baseline is precise, at 82.80\%, but it misses a great deal, at 48.43\% recall. Ours is more balanced, at 79.46\% precision and 92.45\% recall. Table~\ref{tab:confusion} gives the underlying counts. The clearest difference is in missed endpoints, which fall from 82 to 12, a false negative rate of 51.57\% against 7.55\% and close to a sevenfold reduction. We pay for that with more false positives, 38 against 16, an FPR of 23.75\% against 10.00\%. F1 rises from 61.11\% to 85.47\%.

Neither classical baseline comes close, even though both were fitted on the test podcast. The silence threshold scores 49.22\%, no better than chance. The prosody and silence regression reaches 62.07\%, and nearly all of that comes from trailing silence. Fitted on silence alone, the same regression reaches 63.32\%, and on the prosodic features alone, 50.78\%. Section~\ref{sec:discussion} explains why silence behaves this way on BanglaTurn.

\subsection{Ablation study}

Table~\ref{tab:ablation} separates three factors: the encoder, generic or Bangla fine-tuned; the addition of Multi-Scale Pooling (MSP); and the number of unfrozen encoder layers.

With the encoder frozen, the Bangla fine-tuned version edges out the generic one, 84.68 against 83.83 F1. MSP helps once encoder layers are unfrozen, most at 2 unfrozen layers (85.47 against 82.66 F1) and less at 4 (84.75 against 83.29). With a frozen encoder it is slightly worse (84.21 against 84.68). Unfreezing 4 layers with MSP gives the lowest FNR in the table, 5.66\%, but costs precision, 76.92\% against 79.46\% at 2 layers, which suggests the extra capacity starts fitting the training podcasts rather than the task. We settled on the Bangla fine-tuned encoder with MSP and 2 unfrozen layers, at 84.33\% accuracy and 85.47 F1.

These differences should be read with care. The accuracy intervals span about $\pm$4 points, and no two rows differ by more than 3.14 points, so the ablation suggests an ordering of configurations but does not establish that any one is better than another.

\subsection{Quantization analysis}

INT8 dynamic quantization shrinks the model by roughly 73\%, from 148--149 MB to 39--40 MB \cite{jacob2018quantization,krishnamoorthi2018quantizing}, and cuts end-to-end latency by 6--13\%, from \SIrange[range-phrase={--},range-units=single]{180}{191}{\milli\second} to \SIrange[range-phrase={--},range-units=single]{165}{170}{\milli\second}. Table~\ref{tab:quantization} gives the figures per configuration. Accuracy mostly survives quantization. The final model loses 1.26 points and the generic encoder 0.31, but the Bangla fine-tuned encoder without MSP loses 4.70 points, from 84.01\% to 79.31\%.

\section{Discussion}
\label{sec:discussion}

\subsection{Error trade-off and operating point}

The two models fail in opposite directions. Smart-Turn v3 predicts an endpoint for only 93 of the 319 clips and misses 82 of the 159 real ones, so a voice agent built on it would often leave the user waiting in silence. Our model makes the opposite error. At the 0.5 threshold it declares an endpoint on 38 of the 160 clips where the speaker would have carried on, so it would cut in on roughly one such pause in four. Neither balance suits every application. A fast-responding assistant can tolerate an occasional interruption, while cutting the speaker off is costly in dictation or counselling. Because our model outputs a probability, the threshold can be raised to trade missed endpoints for fewer interruptions, or the prediction can be combined with a short silence timeout before the system takes the turn. We report only the 0.5 operating point here, and choosing a threshold on held-out data for a given application is left to future work.

Deployment also calls for checking accuracy after quantization model by model. INT8 costs the final model 1.26 points but costs the Bangla fine-tuned encoder without MSP 4.70 points (Table~\ref{tab:quantization}). We have not isolated the cause of that difference.

\subsection{What silence does and does not tell us}

On BanglaTurn, trailing silence points the wrong way. Endpoint clips end with a median of 0.001\,s of silence against 0.051\,s for non-endpoint clips. That is why a conventional silence threshold scores at chance, while a regression free to learn the reversed direction reaches 63.32\%. The cause is how clips are cut. An endpoint clip ends where diarization detects the next speaker, leaving little silence, while a non-endpoint clip ends at a pause that voice activity detection found inside a turn. This is an artefact of corpus construction, not a property of Bangla turn-taking, and it has two consequences. First, silence duration, the cue deployed systems rely on most, cannot solve the task here, so the benchmark tests other cues. Second, a learned model could use the reversed cue as a shortcut. Our model's 84.33\% is far above the 63.32\% that silence alone reaches, so the shortcut cannot account for its performance, but it may account for part of it. Balancing trailing silence across the two classes in a future release would remove that doubt.

The prosodic features were even less useful on their own, at 50.78\%. The final pitch slope does fall more steeply before endpoints, consistent with the final lowering reported by Ishimoto and Enomoto \cite{ishimoto2016endofutterance} and by Ward and Vega \cite{ward2019prosodic}, but on these short, filler-heavy clips the effect is too weak for a linear model to use. Whatever our model has learned, it is not captured by a handful of utterance-final prosodic measurements.

\subsection{Uncertainty and label quality}

With 319 test clips, every figure carries real uncertainty. The gap to Smart-Turn v3 is large enough to survive it, but most ablation differences are not, and ranking the configurations with confidence would need a larger held-out set covering more podcasts. Label quality adds uncertainty that the intervals do not capture. A single annotator checked every label and also selected the test clips, so there is no agreement figure to bound the label error rate and no independent check on the selection. Where a clip's turn state is genuinely ambiguous, the reference label reflects one person's judgement, and the measured accuracy is partly agreement with that judgement.

\subsection{Beyond podcasts}

All of BanglaTurn is podcast speech, and the 84.33\% figure should not be expected to carry over unchanged to the settings where turn detection matters most commercially. Phone calls are narrowband, typically sampled at 8\,kHz and passed through lossy codecs, which strips much of the spectral detail that a Whisper encoder trained on wideband audio relies on. Voice-assistant interactions are dominated by short commands and questions, whose prosody and pausing differ from the long, planned turns of a podcast conversation. Both settings also bring background noise and overlapping speech that edited podcasts largely avoid. We expect accuracy to drop in each of them, and Bangla test sets built from telephone and assistant speech are the most direct next step.

\section{Conclusions}

We have described the first Bangla turn detection dataset, 35,374 samples built with a pipeline that transfers to other languages, and a Whisper encoder-based model that reaches 84.33\% accuracy, 15.05 points above the baseline in absolute terms and 21.7\% in relative terms. INT8 quantization compresses that model by 73\% and cuts its latency by 13\% at a cost of 1.26 accuracy points. We hope the corpus and the pipeline together give other low-resource languages a place to start.

\subsection{Ethical considerations}
\label{sec:ethics}

We release BanglaTurn through gated access by choice. The audio comes from podcasts that are publicly viewable on YouTube, and although that makes the recordings easy to obtain, we treat them as the work of the people who made them rather than as free material, and we did not seek redistribution permission from individual rights holders. Access is therefore granted on request, for research use only. We release clips of 3 to 15 seconds rather than full episodes, without speaker names or links from clips to their source episodes. Voices are nonetheless identifying and the source recordings remain public, so we do not claim the data is anonymous. Any rights holder or speaker can have their material removed from the dataset by contacting the author at the address on the first page.

\subsection{Limitations}

The dataset covers one domain and one register. Podcast conversation is not phone speech, a voice-assistant exchange, a multi-party meeting or a human-robot exchange, so we cannot claim the model carries over to those settings. The labels rest on a single annotator's check of automatic proposals, with no inter-annotator agreement and no record of how many labels the check changed, so the label error rate is unknown. The same annotator selected the 319 test clips, so the test set is small and its difficulty reflects one person's judgement. Trailing silence differs systematically between endpoint and non-endpoint clips because of how the clips are cut, and a model may exploit it. All error rates are reported at a single decision threshold. The model itself sees a fixed 8-second window, which rules out longer-range discourse cues, works from audio alone, and inherits whatever prosodic information Whisper's features happen to encode.

\section{Generative AI Use Disclosure}

Generative AI was used in two ways. As part of the method, Google's Gemini 2.0 Flash Lite produced first-pass transcriptions and turn-state annotations for the BanglaTurn corpus, as described in Section~\ref{sec:annotation}; the author checked every resulting label by hand. Separately, generative AI assistance was used to edit and polish the wording of this manuscript. It did not produce a significant part of the manuscript, and the author takes full responsibility for the content of the paper.

\bibliographystyle{IEEEtran}
\bibliography{custom}

\end{document}